\documentclass{article} 
\usepackage[table,dvipsnames,svgnames]{xcolor}
\usepackage{iclr2027_conference,times}

\usepackage{amsmath,amsfonts,bm}

\def\eqref#1{equation~\ref{#1}}

\def\1{\bm{1}}

\DeclareMathAlphabet{\mathsfit}{\encodingdefault}{\sfdefault}{m}{sl}
\SetMathAlphabet{\mathsfit}{bold}{\encodingdefault}{\sfdefault}{bx}{n}

\usepackage{array}
\usepackage{multirow}
\usepackage{booktabs}
\usepackage{arydshln} 

\usepackage{graphicx}
\usepackage{fontawesome5}
\usepackage{subcaption}

\definecolor{mvgray}{gray}{0.80}
\definecolor{headercolor}{HTML}{DDEBF7}   
\definecolor{avgcolor}{HTML}{F3F6FA} 

\newcommand{\best}[1]{\textbf{#1}}

\usepackage{hyperref}
\usepackage{url}
\usepackage{cleveref}

\title{MultiLoc: Look Around As You Localize for Fast And Robust Visual Re-localization}

\author{Nobel Dang \& Bing Li\thanks{DISTRIBUTION STATEMENT A. Approved for public release; distribution is unlimited. OPSEC \# 10390}\\
  Clemson University, SC, USA
}

\iclrfinalcopy
\begin{document}

\maketitle

\begin{abstract}

Relative camera pose is a basic geometric cue for visual re-localization and scene understanding. When using images alone, visual evidence may not sufficiently support reliable estimation of 3D-consistent motion. The challenge is especially acute in unfamiliar and ever-changing environments for real-time applications, where a system must infer spatial structure under tight time constraints from what it sees rather than rely on scene-specific reconstruction or training. This raises a fundamental question: how can live 3D camera motion be estimated accurately for re-localization in unseen environments while drawing on enough scene context to resolve pose ambiguity? We introduce MultiLoc, a multi-view-guided relative pose regressor trained at scale to achieve spatial and geometric consistent representations for robust visual re-localization in unseen scenarios with high inference speed. Specifically, MultiLoc creates a minimal 3D-sub-scene representation of the environment and efficiently transforms it into 3D spatially and geometrically consistent features in a single forward pass, enabling precise pose estimates with high inference speed gains. Across diverse indoor, outdoor, and in-the-wild visual re-localization benchmarks---Indoor6, Cambridge Landmarks and WaySpots---MultiLoc consistently outperforms state-of-the-art relative pose regression methods. We also highlight that MultiLoc, a pose regressor, also performs competitively with inference-heavy and structure-based approaches while retaining sub-second inference and generalizing to unseen scenes. Given a small posed support set, it also surpasses relative pose regression, feature-matching, and non-regression methods on several relative camera pose benchmarks. Code will be released.

\end{abstract}

\section{Introduction}
\label{sec:intro}

Visual Re-localization estimates an agent's position and orientation within an environment from visual observations alone. The said environment is represented by reference images and their camera poses. The poses, obtained through Structure-from-Motion, 3D scanners, or geo-tagging, provide a global anchor that corrects drift inherent in odometry and incremental SLAM. It also supports accurate pose estimation, system initialization, and loop closure, especially in GNSS-denied environments. Such capabilities make visual re-localization important for autonomous driving \cite{liu2021visual}, navigation \cite{suomela2023benchmarking}, augmented reality \cite{castle2008video}, and other robotic applications.

\begin{figure}[ht]
  \centering
   \includegraphics[width=0.98\textwidth]{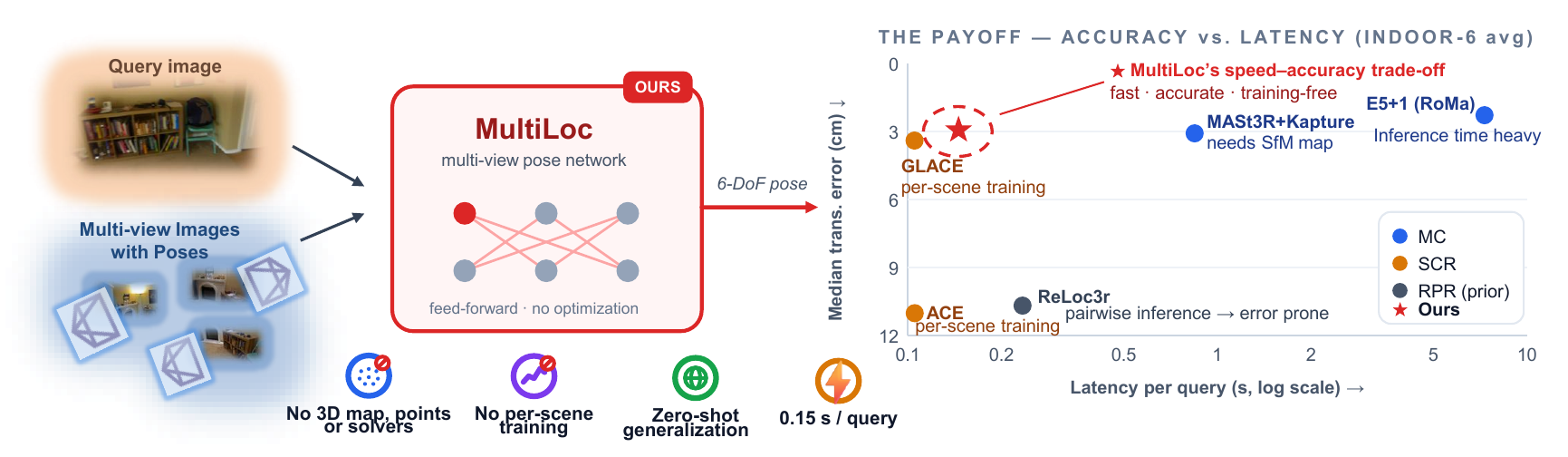}
   \caption{\textbf{One forward pass, no scene-training or map.} Given a query and $K$ reference images with 6-DoF poses---a camera constellation, or 3D sub-scene (left)---MultiLoc achieves sub-scene consistent spatial and geometric representations by jointly fusing multi-view images and the available poses. By doing so, just with a single forward pass, MultiLoc matches the accuracy of feature matching (MC) and scene-coordinate regression (SCR) methods while offering one of the fastest inference speed across all the baselines on Indoor-6 (right).}
    \label{fig:teaser} 
   \vspace{-4mm}
\end{figure}

Traditionally, structure-based methods ~\cite{sarlin2021back, sattler2016efficient, taira2018inloc} build a 3D map, match image features to map points, and estimate camera pose with PnP \cite{gao2003complete} and RANSAC ~\cite{fischler1981random}. They require high mapping time, substantial descriptor storage, and slow inference. Recent 3D foundation models like MASt3R \cite{leroy2024grounding} achieve state-of-the-art re-localization without storing descriptors but they still require substantial mapping time and memory for Structure-from-Motion (SfM) reconstruction ~\cite{bruns2025ace}, while remaining significantly slower during inference than relative pose regression (RPR) methods. Conversely, scene coordinate regression (SCR) methods \cite{brachmann2021visual, li2020hierarchical, nguyen2024focustune, wang2024hscnet++, brachmann2023accelerated} implicitly encode scenes to directly predict 2D-3D correspondences. Recent variants like the ACE family ~\cite{brachmann2023accelerated, wang2024glace, bruns2025ace} even enhance mapping efficiency, scene scalability, and robustness to domain shifts. However, both paradigms fundamentally necessitate per-scene optimization and an inference-time pose solver.

Interestingly, a third line of work under structureless feature-matching forgoes the 3D model entirely. Semi-generalized E5+1 solver \cite{zheng2015structure, panek2026guide}, for example, determines relative pose using five correspondences to a primary retrieved image and resolves absolute scale via one correspondence to a second, bypassing both explicit 3D maps and per-scene training. However, such methods typically operate on pair of images for 2D-2D matches and rely on geometric solvers while trading off their accuracy with inference speed compared to pose regression.

For real-time applications, pose regression methods are generally sought that encode scene representation in network weights and predict camera pose. Relative Pose Regression (RPR) methods, such as ~\cite{dong2025reloc3r, li2025geloc3r, chen2021wide, balntas2018relocnet} are trained on image pairs to predict the relative camera pose between the two images. For visual localization, during inference, one image is the query image while another typically comes from the top-k retrieved geo-tagged reference images of the environment. The predicted relative poses are then utilized with ground-truth poses of the reference (or database) images to recover the absolute pose using methods like Umeyama alignment ~\cite{umeyama2002least} or motion-averaging module ~\cite{dong2025reloc3r}. Recently, ~\cite{dong2025reloc3r, li2025geloc3r} has shown the generalizability and faster inference speed capabilities of such feed-forward RPR based methods when trained on a large-scale of data. Despite achieving high performance and inference speed, pose regression methods often exhibit inferior performance compared to other paradigms. We attribute this limitation to a lack of global spatial awareness; by training on image pairs, these models merely approximate poses from pairwise relationships ~\cite{arnold2022map} rather than capturing the broader scene context. Furthermore, such works typically estimate relative camera poses without 3D-aware constraints, neglecting the geometric and structural consistency between pairs.

To address these limitations, we introduce \textbf{MultiLoc}, a novel large-scale trained relative pose regressor. It comprises of a multi-view guided pose regressor that aggregates holistic spatial cues from $k$ reference images and their corresponding ground-truth camera poses in a single forward pass. Unlike existing RPR methods that rely solely on pairwise visual appearance, MultiLoc integrates multi-view inputs and poses to achieve superior spatial and geometric grounding as shown in \Cref{fig:teaser}. By incorporating multiple reference views and poses directly into the regression phase, our model anchors the query image within a known 3D context, or what we refer as \textit{3D sub-scene context}, providing highly accurate estimates for the camera pose estimation task. For re-localization, we follow the two stage pipeline that includes image retrieval followed by our pose regressor. It is worth noting that our work for re-localization is independent of any camera calibration parameters, scene-specific training and post-hoc geometric solver or optimization for metric scale recovery. Instead, MultiLoc simply \textit{looks around} in the 3D sub-scene—using visual observations and their camera poses to estimate the query's pose precisely. Thus, MultiLoc departs from existing RPR methods, which regress relative poses from isolated image pairs and afterwards attain global metric poses through geometric solvers that inherit each pairwise estimate's errors. Finally, we demonstrate the zero-shot capability of our proposed method, MultiLoc, by benchmarking  on multiple datasets. To the best of our knowledge, MultiLoc is the first pose regression work to achieve visual re-localization performance comparable to other non-regression methods like scene coordinate regression or feature matching.
Therefore, the main contributions of this work are as follows:

\begin{enumerate}

\item We propose a new approach for relative pose regression by recasting it to \emph{anchor the query in a 3D sub-scene context during estimation for re-localization.} This helps grounding query pose prediction with multi-view spatial and geometric consistent features.

\item For effective 3D-sub-scene representation learning and grounding the query in such, we introduce a novel multi-view and geometry-aware relative pose regressor.

\item We present MultiLoc, a two-stage re-localization framework, based on our proposed regressor, attaining robust re-localization performance in unseen environments with only one forward pass and real-time inference speed without scene-specific optimizations.

\item Finally, we investigate effectiveness of the newly proposed paradigm by benchmarking on relative camera pose estimation, where our regressor outperforms both non-pose-regression and pose regression approaches; and on visual re-localization where MultiLoc achieves state-of-the-art within RPR and comparable performance with non-regression methods.
\end{enumerate}

\section{Related Work}
\label{sec:relatedwork}
Typically, a common body of work addresses both relative pose estimation and visual re-localization:

\noindent\textbf{Structure-based Methods.} These methods estimate pose from 2D–3D correspondences and typically require scene-specific mapping or training as discussed in \Cref{sec:intro}, though ACE variants train in minutes~\cite{brachmann2023accelerated, wang2024glace, bruns2025ace}. Scene-agnostic regressors~\cite{revaud2024sacreg, tang2021learning} avoid retraining but add inference costs through cost volumes or sparse SfM correspondences. Most recently, FastForward~\cite{barroso2026scene} reduces mapping to retrieval and localizes in one feed-forward pass using ray-encoded features sampled from posed images instead of an explicit 3D model. However, it still regresses scene coordinates and uses PnP-RANSAC, requiring known camera intrinsics. MultiLoc directly regresses pose and avoids these requirements. Although FastForward advances visual re-localization, its retrieval GeM-AP checkpoint and source code are unavailable, preventing a fair direct benchmark. Nonetheless, we report its performance in the appendix for completeness.

\noindent\textbf{Pose Regression Methods.} Such regressors forgo intermediate correspondences and predict pose directly, either in a scene's absolute frame, called Absolute Pose Regression (APR), ~\cite{kendall2015posenet, brahmbhatt2018geometry, shavit2021learning, chen2021direct} or relative to another image, termed as Relative Pose Regression (RPR), ~\cite{dong2025reloc3r, li2025geloc3r, balntas2018relocnet, chen2021wide, zhou2020learn, winkelbauer2021learning}; both offer real-time inference, with APR trailing RPR in cross-scene generalization. Bridging APR and structure-based methods, marepo ~\cite{chen2024map} decouples pose regression into a per-scene geometry network that predicts scene coordinates and a scene-agnostic transformer that regresses pose directly from them, attaining accuracy competitive with structure-based methods while replacing the explicit solver with a learned forward pass; its geometry network, however, is per-scene trained, so marepo still incurs mapping time, unlike the fully feed-forward RPR paradigm we adopt. Map-free re-localization \cite{arnold2022map} instead bridges RPR and feature matching, regressing metric relative pose to a single reference image via monocular depth and explicit matching, though it depends on known camera intrinsics for geometric back-projection. MultiLoc is likewise an RPR method, combining high accuracy and inference speed with no scene-specific training; accordingly, our localization benchmarks focus on comparison against such methods.

\noindent\textbf{3D Foundation Models.} Recent works use large-scale pre-training to learn generalizable 3D representations. DUSt3R ~\cite{wang2024dust3r}, building upon CroCo \cite{weinzaepfel2023croco}, pioneered this shift with a pointmap representation that establishes dense 2D-3D correspondences from a pair of images. CUT3R~\cite{wang2025continuous} extends this approach to image streams through recurrent state updates. To reason over all views jointly, VGGT \cite{wang2025vggt} introduced a unified transformer that processes multiple images simultaneously, capturing richer spatial context than independent pairs. A common limitation of the image-only variants is their inability to exploit auxiliary priors---such as known camera poses---that are readily available in visual re-localization and can offer tight geometric constraints. Pow3R \cite{jang2025pow3r} and G-CUT3R \cite{khafizov2025g} address this for local paradigms: Pow3R conditions DUSt3R on camera intrinsics, poses, or depth, while G-CUT3R injects the same modalities into CUT3R's recurrent decoder via zero-convolution. However, both remain bounded by \emph{local} processing---pairwise for Pow3R, sequential for G-CUT3R---so scene-level consistency across many views must still be recovered through post-hoc alignment or accumulated in a compressed state prone to drift. Concurrent to our work, MapAnything \cite{keetha2026mapanything} generalizes this joint paradigm into a universal reconstruction backbone that accepts a variable number of images and optional geometric inputs such as camera intrinsics, poses, or depth. Therefore, we include MapAnything as a strong baseline in our work.

\vspace{0.5em}
\noindent\textbf{Visual Re-Localization.} Building on these foundation models, ReLoc3r \cite{dong2025reloc3r} specialized DUSt3R for visual localization, achieving state-of-the-art relative pose regression through large-scale supervision. Re-localization, however, has a specific structure that these methods do not fully exploit. The task pairs a query image with a set of reference frames whose camera poses are known. We therefore propose to leverage a joint multi-view transformer, as in VGGT \cite{wang2025vggt}, conditioned on the poses of the reference images. We hypothesize that jointly processing all available views together with the reference camera poses facilitates a more holistic and better scene-grounded understanding of spatial context, leading to more robust pose estimation.

\section{Method}
\label{sec:method}
\noindent\textbf{Problem Statement.} Given a database of \textit{k} images, $\mathbf{I} = \{I_i\}_{i=1}^k, \mid I_i \in \mathbb{R}^{H \times W \times 3}$, from a scene with their absolute camera poses, $P = \{p_1, p_2, \dots, p_k\} \mid p_i \in \text{SE}(3)$, in a common world coordinate system, visual re-localization aims to find the absolute camera pose of a query image $I_q$.

\noindent\textbf{Method Overview.}
We present MultiLoc, a robust relative pose regression framework designed for visual re-localization and camera pose estimation task. Unlike existing methods such as \cite{dong2025reloc3r, li2025geloc3r}, which operate on isolated image pairs, MultiLoc ingests a sequence of $(k+1)$ images. By utilizing the first $k$ relative camera poses as geometric priors along with the images, our model achieves a more comprehensive spatial understanding of the scene, leading to significantly enhanced relative camera pose estimation accuracy for the query frame.

\begin{figure*}[t]
  \centering
   \includegraphics[width=0.82\textwidth]{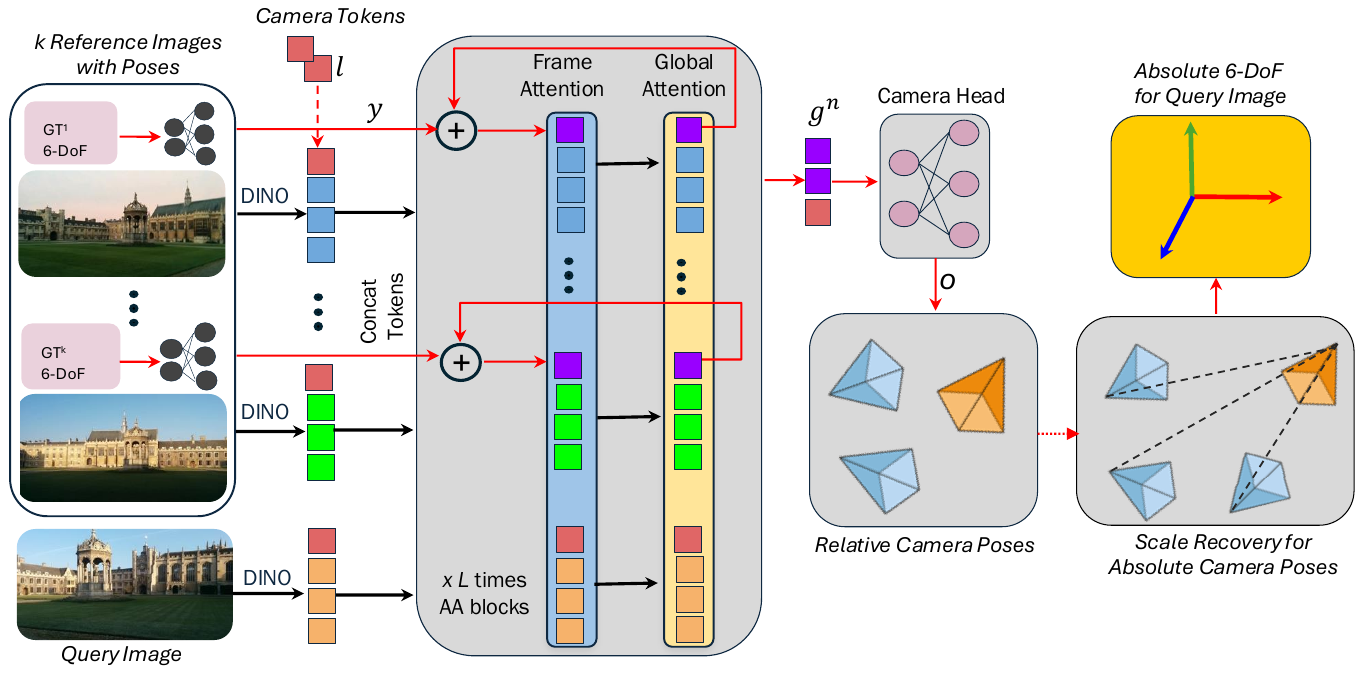}
   \caption{\textbf{MultiLoc Architecture overview.} The retrieved images’ extrinsics are embedded via our camera encoders into camera tokens $y$ and added with learnable camera tokens $l$, while all images pass through DINO-v2 to obtain patch tokens. All the tokens are then concatenated and processed by alternating-attention (AA) transformer blocks. Finally, camera tokens are used to predict relative poses between query and reference images using the camera head and metric scale is recovered by the pre-computed scene-scale $s$. The \textcolor{red}{red} arrows denote the flow of camera tokens.}
   \label{fig:architecture}
   \vspace{-6mm}
\end{figure*}

\subsection{Architecture}
\label{subsec:arch}
We begin by describing the architecture that is common for both camera pose estimation and visual re-localization and as shown in \Cref{fig:architecture}. Input to our model at a time are \textit{k+1} images and \textit{k} corresponding relative camera poses. The poses are input as members of $SE(3)$ manifold which are flattened to $\mathbb{R}^{12}$. A small multi-layer perceptron (MLP) is then applied to encode each of the camera pose as:
$\text{MLP} : p_i \in \mathbb{R}^{12} \mapsto y_i \in \mathbb{R}^{d}$. The poses are normalized by the average distance, $s = \frac{1}{k} \sum_{i=1}^{k} \|c_i\|$ where $c_i$ is the translation component of a camera pose.

\noindent The images, $I = \{I_1, \dots , I_{k}, I_q\}$, are tokenized in a patch of size 14 and fed into the DINO-v2 \cite{oquab2023dinov2} model to obtain per-patch features $f \in \mathbb{R}^{d}$. Learnable camera tokens, $l=\{l_i\}_{i=1}^{k+1}$ are then added with ground-truth camera tokens, $y$, to form final pose tokens as:
$g = l + y$. For the query image, we use $y_q = \mathbf{0} \in \mathbf{R}^d$. Finally, pose tokens are concatenated with register tokens, $r$ and image patch tokens, $f$, to form the input to our transformer block as follows: $t = [g, r, f],$ 
where [.] is a concatenation operation.

Finally, our transformer model ingests all the tokens and first perform self (or frame) attention followed by cross (or global) attention across all the tokens $t$ similar to \cite{wang2025vggt}. This allows each token, especially query image tokens, to be grounded with scene's 3D spatial context which pair-wise peers lag. The transformer block is repeated $L$ times where after every frame attention we add the ground truth encoded pose tokens, $y$, for long-term context as $g^n = g^o + y$ where $g^o$ is the output from previous layer and $g^o = g$ for the first transformer block.

\noindent \textbf{Camera Pose Estimation.} The learnable camera tokens are sampled from the $L-th$ transformer block's output and input into a simple four-layer self-attention-based camera head followed by a linear layer. The camera head predicts both the relative camera extrinsic and intrinsic parameters as:
$$C(.): g^n \mapsto o \in [q, c, f_x, f_y]$$ where $q_i$ is the quaternion representation $\in \mathbb{R}^4$, $c_i \in \mathbb{R}^3$ is the translation component at a normalized scale and finally $f_x$ and $f_y$ are the focal lengths.

\subsection{Visual Re-Localization}
\label{subsec:visualloc}
We retrieve the top-10 reference images using NetVLAD \cite{arandjelovic2016netvlad} and process them with the query as described in \cref{subsec:arch}. The query camera token predicts $o_{1,q}$ in the first reference frame at normalized scale. Rescaling its translation by $s$ and composing with the first reference pose yields the global metric pose without geometric optimization. We compare this direct recovery with Umeyama alignment and motion averaging in the appendix \Cref{tab:scale_recovery}.

\subsection{Training Details}
We initialize MultiLoc's weight with VGGT ~\cite{wang2025vggt} and finetune on 4 H200s with a total of 1M iterations. Our model comprises of 1.1 billion parameters with $L=24$ total transformer blocks and is trained using AdamW with learning rate of $1 \times 10^{-5}$ decaying to $1 \times 10^{-8}$. The aspect ratio of input images is randomized to be between 0.33 and 1, where the width is resized to be of maximum length as 518 pixels. Data augmentations and input image sampling is applied similar to VGGT. For training we utilize Co3D ~\cite{reizenstein2021common}, DL3DV ~\cite{ling2024dl3dv}, VKITTI2 ~\cite{cabon2020virtual}, BlendedMVS~\cite{yao2020blendedmvs}, Re10K ~\cite{zhou2018stereo}, MegaDepth ~\cite{li2018megadepth}, ARKitScenes ~\cite{baruch2021arkitscenes} and ScanNet++ ~\cite{yeshwanth2023scannet++}.

\noindent\textbf{Loss function.}
For training we extend the homoscedastic uncertainty-based geometric loss function \cite{kendall2018multi} to regress on both the camera extrinsic and intrinsics parameters as follows:

\noindent
\begin{minipage}{0.42\textwidth}
\begin{equation}
    \label{eq:loss}
    \text{loss}_{x} = |x^{\text{pred}} - x^{\text{gt}}|
\end{equation}
\end{minipage}
\hfill
\begin{minipage}{0.55\textwidth}
\begin{equation}
    \label{eq:total_loss}
    \text{loss}(.) = \sum_{x} {\text{loss}_{\text{x}} \cdot e^{-s_x} + s_x}
\end{equation}
\end{minipage}

where $x$ is a variable being translation, rotation or focal length component and $s_x \in \{s_c, s_q, s_f\}$ are the learnable homoscedastic uncertainty scalar values for each of the $x$ task variables respectively.

\section{Experiments}
\label{sec:experiment}
\noindent \textbf{Setup.}
Given that MultiLoc is a large-scale trained model, we evaluate its zero-shot generalization across diverse environments, including indoor, outdoor, natural and in the-wild domains. Our benchmarking focuses on two primary tasks: relative camera pose estimation and visual re-localization. We report AUC@$\theta$ for relative pose estimation following \cite{dong2025reloc3r}, and median translation error (m or cm) and median rotation error ($^\circ$) for visual re-localization.

For both evaluation tasks, we benchmark MultiLoc against representative methods from two broad categories: pose-regression and non-pose-regression approaches, including feature matching and scene-coordinate regression for comparative analysis  on visual re-localization. However, it should be noted that MultiLoc is a relative pose regression method.

\subsection{\textbf{Relative Camera Pose Estimation}}
\label{subsec:rel_cam_pose}

\begin{table*}[t]
\centering
\caption{
Relative camera pose estimation results on ACID, MegaDepth-1500, and ScanNet-1500.
\textbf{Non-PR} denotes non-pose-regression methods, and \textbf{PR} relative pose regression. Shaded rows use multiple views; unshaded rows are pairwise. \textcolor{red}{Red} indicates the best and \textbf{bold} the second-best result.
}
\label{tab:results_rcp}

\setlength{\tabcolsep}{8pt}
\resizebox{\textwidth}{!}{%
\begin{tabular}{ccl ccc ccc ccc}
\toprule
& & \multirow{2}{*}{\textbf{Method}}
& \multicolumn{3}{c}{\textbf{ACID}}
& \multicolumn{3}{c}{\textbf{MegaDepth-1500}}
& \multicolumn{3}{c}{\textbf{ScanNet-1500}} \\

\cmidrule(lr){4-6}
\cmidrule(lr){7-9}
\cmidrule(lr){10-12}

& &
& $5^\circ$ & $10^\circ$ & $20^\circ$
& $5^\circ$ & $10^\circ$ & $20^\circ$
& $5^\circ$ & $10^\circ$ & $20^\circ$ \\
\midrule

& & Efficient LoFTR 
& -- & -- & --
& 52.80 & 69.19 & 81.18
& 19.20 & 37.00 & 53.60 \\

& & RoMa
& 57.98 & 68.53 & 76.62
& 60.92 & 75.63 & 85.69
& 34.13 & 55.28 & 71.69 \\

& & DUSt3R
& 21.50 & 35.95 & 49.70
& 27.90 & 46.00 & 63.30
& 23.81 & 45.91 & 65.57 \\

& & MASt3R (Ess.)
& 54.42 & 66.23 & 75.26
& 42.79 & 62.55 & 77.61
& 35.25 & 57.64 & 74.48 \\

& \multirow{-5}{*}{\rotatebox[origin=c]{90}{\scriptsize Pair}}
& NoPoSplat
& 48.60 & 61.70 & 72.80
& -- & -- & --
& 31.80 & 53.80 & 71.70 \\

\cdashline{2-12}

\rowcolor{mvgray}
& & MASt3R (Ess.)
& 49.51 & 64.68 & 76.08
& 48.68 & 68.16 & 82.20
& 33.40 & 55.27 & 73.42 \\

\rowcolor{mvgray}
\multirow{-7}{*}{\rotatebox[origin=c]{90}{\textbf{Non-PR}}}
& \multirow{-2}{*}{\rotatebox[origin=c]{90}{\scriptsize MV}}
& RoMa
& 58.54 & 70.10 & 78.79
& 68.05 & \textbf{81.23} & \textcolor{red}{89.76}
& 28.22 & 49.91 & 68.88 \\

\midrule

& & ReLoc3r 
& 38.18 & 56.39 & 70.34
& 49.59 & 67.84 & 81.22
& 34.78 & 58.31 & 75.52 \\

& & VGGT 
& 22.59 & 40.83 & 59.04
& 40.93 & 60.81 & 76.61
& 37.10 & 58.93 & 75.51 \\

& & MapAnything 
& 26.31 & 45.95 & 63.46
& 27.17 & 47.83 & 67.04
& 24.65 & 47.69 & 67.64 \\

& \multirow{-4}{*}{\rotatebox[origin=c]{90}{\scriptsize Pair}}
& MultiLoc (Ours)
& 23.83 & 40.70 & 58.08
& 46.64 & 63.16 & 75.82
& 32.62 & 55.03 & 72.65 \\

\cdashline{2-12}

\rowcolor{mvgray}
& & VGGT 
& 22.55 & 40.85 & 58.75
& 46.25 & 66.46 & 80.70
& 38.82 & 60.31 & 76.48 \\

\rowcolor{mvgray}
& & MapAnything 
& 37.24 & 57.50 & 73.34
& 34.45 & 54.82 & 72.08
& 34.97 & 56.77 & 73.62 \\

\rowcolor{mvgray}
\multirow{-7}{*}{\rotatebox[origin=c]{90}{\textbf{PR}}}
& \multirow{-3}{*}{\rotatebox[origin=c]{90}{\scriptsize MV}}
& \textbf{MultiLoc (Ours)}
& \textcolor{red}{76.56}
& \textcolor{red}{87.15}
& \textcolor{red}{93.31}
& \textcolor{red}{69.49}
& \textcolor{red}{81.98}
& \textbf{89.61}
& \textcolor{red}{62.34}
& \textcolor{red}{76.32}
& \textcolor{red}{86.57} \\

\bottomrule
\end{tabular}%
}
\vspace{-4mm}
\end{table*}

\noindent Relative camera pose estimation recovers relative pose between images. We evaluate pairwise relative pose estimation on ScanNet-1500~\cite{dai2017scannet, sarlin2020superglue}, MegaDepth-1500~\cite{sun2021loftr}, and the Aerial Coastline Imagery Dataset (ACID)~\cite{liu2021infinite}. The datasets cover complementary settings, including indoor scenes, crowd-sourced outdoor landmark photos with diverse viewpoints, and sequential drone footage of natural coastlines. MegaDepth-1500 is drawn from the same source as our training MegaDepth. ScanNet-1500 scenes are sampled from Scannet and are completely disjoint our training on ScanNet++ while ACID's domain is entirely unseen.

\textbf{Baselines.} We categorize baselines into: \emph{Pose Regression} (PR) including state-of-the-art ReLoc3r~\cite{dong2025reloc3r} along with feed-forward multi-view models VGGT~\cite{wang2025vggt} and MapAnything~\cite{keetha2026mapanything}. And \emph{Non-Pose-Regression} (Non-PR) with Eff. LoFTR~\cite{wang2024efficient}, RoMa~\cite{edstedt2024roma} as feature matchers along with DUSt3R~\cite{wang2024dust3r}, MASt3R~\cite{leroy2024grounding}, and the pose-free NoPoSplat~\cite{ye2024no}.

Since, MultiLoc is designed to be guided by additional views, we evaluate every method in two regimes. In the \textbf{Pairwise (Pair)} regime the model sees only the intended pair $(I_q, I_t)$. In the \textbf{Multi-view (MV)} regime we use $k$=10 support views, randomly sampled per pair, together with their camera poses. The AUC is always measured on the intended pair $(I_q, I_t)$ alone, and every MV method — pose and non-pose regression alike — is given the \emph{same} support set. We additionally report MultiLoc and its multi-view counterparts in the Pairwise regime for completeness. Among the pose regressors, MapAnything, like MultiLoc, ingests both the support images and their poses, whereas VGGT uses the images alone. To extend the same advantage to the Non-PR methods and to verify that any gain is not merely due to the smaller viewpoint changes afforded by support views, we also supply them the support set. For the matchers (RoMa, MASt3R) we run matching on the intended pair $(I_q, I_t)$ and on each query–support pair $(I_q, I_{s_i})$, select the pair with the highest inlier count, and report the performance using query and the selected view. Although all pairs are candidates, the oracle selection can underperform Pairwise because higher inlier count does not guarantee better pose accuracy, see appendix \Cref{fig:roma_multiview} and \Cref{subsec:multiview_matching}.

\textbf{Conclusion.} As shown in \Cref{tab:results_rcp}, MultiLoc is competitive with dedicated regressors in Pairwise mode and achieves the best MV AUC on ACID and ScanNet-1500, and at $5^\circ/10^\circ$ on MegaDepth-1500. It substantially outperforms MapAnything with identical posed support, while VGGT gains little from the same multi-view images. MultiLoc also surpasses the feature-matching baselines that utilize the same additional support set. These results highlight that grounding query in a posed 3D sub-scene induces geometrically and sub-scene–consistent features, enabling better pose accuracy.

\subsection{\textbf{Visual Re-Localization}}
\label{localization}
We evaluate MultiLoc for visual re-localization against methods drawn from three families for comparative benchmarking. First is our own \emph{relative pose regression} (RPR) family. Second is \emph{matching-based} methods (MC) which localize through feature matching, in either a 2D--3D or a 2D--2D fashion, and, finally, \emph{scene coordinate regression} (SCR) based methods. Alongside performance metrics, we report per-query latency on a single H100 GPU and mapping time. For SCR methods, latency is dominated by the CPU-side RANSAC/PnP solver, so it varies with scene difficulty, implementation, and hardware. We therefore, for SCR, treat absolute latency as indicative, while relative comparisons remain meaningful. Furthermore, mapping time depends on factors such as scene size. Therefore, we report bounds using all database images, without subsampling.

\noindent\textbf{Baselines.} Within the matching-based family we include two complementary approaches. MASt3R+Kapture \cite{leroy2024grounding} establishes 2D--3D correspondences between the query and a pre-built 3D map and recovers the pose with a PnP solver. E5+1 \cite{panek2026guide} instead relies purely on 2D--2D correspondences between the query and a set of posed reference images, obtained with the dense matcher RoMa \cite{edstedt2024roma}. For scene coordinate regression we benchmark state-of-the-art ACE \cite{brachmann2023accelerated}, and its large-scale extension GLACE \cite{wang2024glace}. Finally, within our own RPR family, we compare against state-of-the-art ReLoc3r \cite{dong2025reloc3r} and GeLoc3r \cite{li2025geloc3r}, wherever possible since it's code is not public yet. Please note all the methods relying on image retrieval are provided with the same references. Implementation details for all baselines are mentioned in appendix.

\begin{figure}[ht]
  \centering
  \begin{subfigure}[b]{0.39\textwidth}
    \centering
    \includegraphics[width=\linewidth]{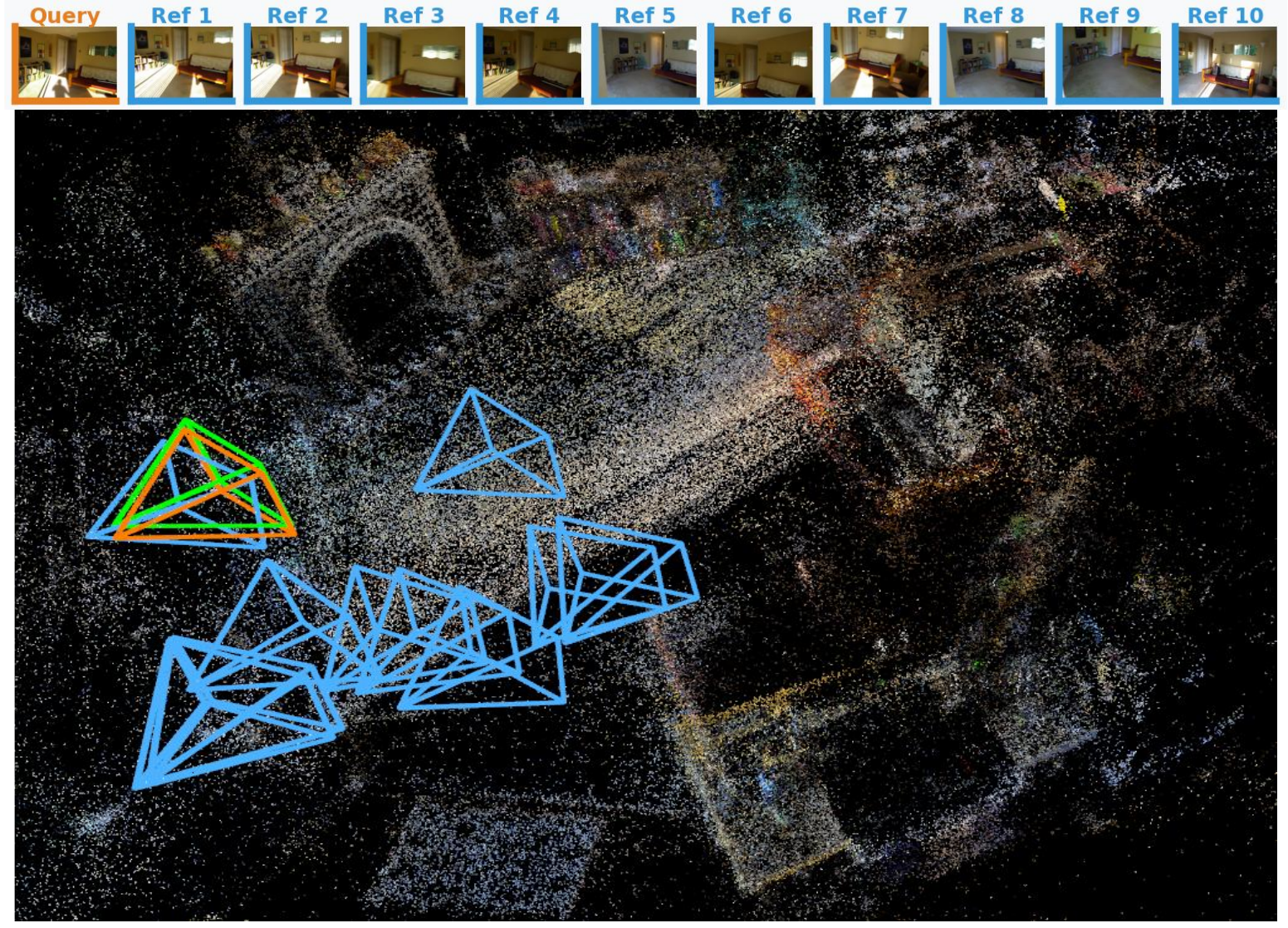}
    \caption{}
    \label{fig:indoor6_loc}
  \end{subfigure}
  \hfill
  \begin{subfigure}[b]{0.48\textwidth}
    \centering
    \setlength{\tabcolsep}{5pt}
    \renewcommand{\arraystretch}{1.25}
    \resizebox{\linewidth}{!}{%
    \begin{tabular}{ll c c}
    \toprule
    \rowcolor{headercolor}
    \textbf{Categ.} & \textbf{Method} & \textbf{Avg (cm/$^\circ$)} & \textbf{Lat.} \\
    \midrule
    \multirow{2}{*}{\textit{MC}}
     & MASt3R + Kapture & 3.1 / \best{0.30} & \cellcolor{blue!20}0.85 s \\
     & E5+1 (RoMa)      & \best{2.3} / 0.45 & \cellcolor{red!20}7.30 s \\
    \midrule
    \multirow{2}{*}{\textit{SCR}}
     & ACE              & 11.0 / 1.80 & \cellcolor{green!20}$\sim$0.10 s \\
     & GLACE            & \best{3.5} / \best{0.60} & \cellcolor{green!20}$\sim$0.10 s \\
    \midrule
    \multirow{2}{*}{\textit{RPR}}
     & ReLoc3r          & 10.7 / 0.80 & \cellcolor{green!20}0.23 s \\
     & MultiLoc (ours)  & \best{2.9} / \best{0.43} & \cellcolor{green!20}0.15 s \\
    \bottomrule
    \end{tabular}%
    }
    \caption{}
    \label{tab:indoor6_compile_results}
  \end{subfigure}
  \caption{\textbf{Visual re-localization on the Indoor-6 benchmark.}
  \textbf{(a)} Top row shows the \textcolor{Orange}{query} and retrieved \textcolor{cyan}{reference images}. Bottom row shows the Indoor6 reconstruction, where MultiLoc's \textcolor{LimeGreen}{predicted pose} closely matches the \textcolor{Orange}{ground truth} in position and orientation. \textcolor{cyan}{Blue frustums} indicate reference poses.
  \textbf{(b)} Median translation\,/\,rotation error averaged over six scenes (cm\,/\,$^\circ$; lower is better) and per-query latency; \textbf{bold} is best within each category. Per-scene results are in appendix.}
  \label{fig:indoor6}
\end{figure}

\subsubsection{\textbf{Indoor Visual Re-Localization.}}
To benchmark cross-dataset transferability of our method, we conduct zero-shot evaluations on Indoor6 \cite{do2022learning} dataset. Unlike static benchmarks, Indoor6 introduces dynamic environmental factors, including day-night transitions and illumination changes across six diverse scenes. \Cref{tab:indoor6_results} from appendix and \Cref{tab:indoor6_compile_results} highlights the efficacy of MultiLoc, where it achieves competitive state-of-the-art (SOTA) performance, validating its ability to handle the geometric and photometric complexities of indoor scenes without prior fine-tuning. \Cref{fig:indoor6_loc} demonstrates the qualitative result of visual re-localization performance of MultiLoc on Indoor6 benchmark. MultiLoc outperforms existing relative pose regressors, reducing median translation and rotation errors by $72\%$ and $50\%$, respectively. Remarkably, it surpasses both of the scene coordinate regression (SCR) methods and a map-based matching approach, MASt3R+Kapture, without relying on explicit 3D reconstructions.

\subsubsection{\textbf{Outdoor and In-the-wild Visual Re-Localization.}}

\begin{figure*}[h]
  \centering
   \includegraphics[width=0.85\textwidth]{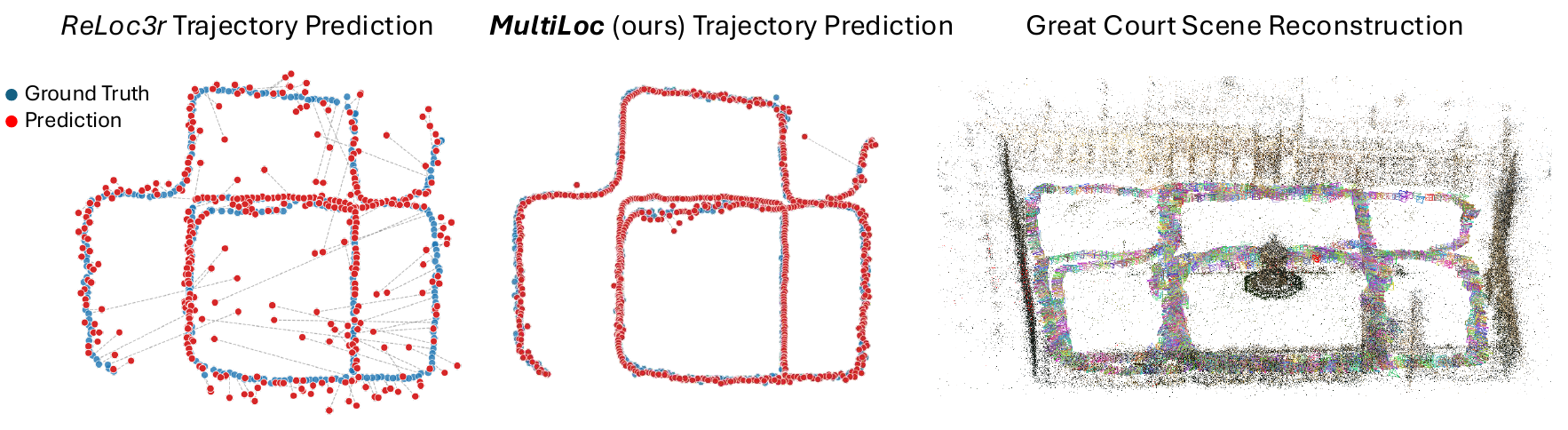}
   \caption{\textbf{MultiLoc v/s ReLoc3r trajectory error} depiction on GreatCourt sequence of Cambridge dataset. Gray dashed lines indicate the translational error between \textcolor{cyan}{ground truth} and \textcolor{red}{predicted} poses. MultiLoc achieves significantly lower error by leveraging 3D sub-scene context from multiple images and their pose, whereas ReLoc3r’s reliance on pairwise images lead to higher drift.}
   \label{fig:trajectory}
   \vspace{-4mm}
\end{figure*}

\noindent We evaluate MultiLoc on outdoor Cambridge Landmarks \cite{kendall2015posenet} and in-the-wild WaySpots \cite{brachmann2023accelerated} benchmarks. Cambridge Landmarks dataset provides diverse scenes characterized by significant environmental clutter and temporal variations in lighting and weather. On the other hand, WaySpots dataset, incorporates both small places of interests like lawn or sculptures and challenging environments like wintersign. Again, both of these datasets are entirely not seen by our model. 
Our method significantly enhances re-localization performance, yielding improvements of around 43\% and 38\% on the Cambridge , see \Cref{tab:cambridge}, and 80\% and 72\% on WaySpots benchmark for translation and rotation, respectively in the RPR family, \Cref{tab:wayspots}. The robustness of our method is also qualitatively validated in \Cref{fig:trajectory} by comparing the trajectory prediction.

Across both benchmarks, MultiLoc is competitive with \textit{MC} and \textit{SCR} methods. On Cambridge, its translation error is 0.29 m, comparable to E5+1 (0.21 m) and ACE (0.26 m). Rotation errors among state-of-the-art methods differ by at most 0.21$^\circ$, making fine-grained comparisons less meaningful. On WaySpots, MultiLoc achieves the lowest translation error across method families, while its rotation error is within 0.1$^\circ$ of E5+1. These results suggest that aggregating multi-view images and camera poses into a shared scene representation helps the model learn geometry-aware, spatially consistent features and resolve pose ambiguity.

\begin{table}[h]
\centering
\caption{Visual re-localization benchmark on Cambridge Landmarks with median errors as (m~/~$^\circ$). Green, blue, and red cells indicate the best, middle, and worst comparative results, respectively.}
\label{tab:cambridge}
\setlength{\tabcolsep}{7pt}
\renewcommand{\arraystretch}{1.3}
\resizebox{\textwidth}{!}{%
\begin{tabular}{ll ccccc | c | c | c}
\toprule
\rowcolor{headercolor}
\textbf{Category} & \textbf{Method} & \textbf{GreatCourt} & \textbf{Kings} & \textbf{OldHosp} & \textbf{ShopFac} & \textbf{StMary} & \textbf{Avg} & \textbf{Map. Time} & \textbf{Latency} \\
\midrule

\multirow{2}{*}{\textit{MC}}

 & MASt3R + Kapture
 & \best{0.10} / \best{0.10} & \best{0.08} / \best{0.12} & \best{0.13} / \best{0.24}
 & \best{0.03} / \best{0.13} & \best{0.04} / \best{0.12} & \best{0.08} / \best{0.14} & \cellcolor{red!20}{$>10 min$} & \cellcolor{blue!20} 0.85 s \\
 & E5+1 (RoMa)
 & 0.37 / 0.16 & 0.21 / 0.35 & 0.29 / 0.64
 & 0.07 / 0.27 & 0.11 / 0.37 & 0.21 / 0.36 & \cellcolor{green!20}{$< 1min$} & \cellcolor{red!20} 7.30 s \\
\midrule

\multirow{2}{*}{\textit{SCR}}
 & ACE
 & 0.44 / 0.20 & 0.31 / 0.40 & 0.30 / 0.60
 & 0.06 / 0.30 & 0.20 / 0.60 & 0.26 / 0.42 & \cellcolor{blue!20}{$>2 min$} & \cellcolor{green!20} $\sim$0.10 s \\
 & GLACE
 & \best{0.19} / \best{0.10} & \best{0.19} / \best{0.30} & \best{0.18} / \best{0.40}
 & \best{0.05} / \best{0.20} & \best{0.10} / \best{0.30} & \best{0.14} / \best{0.26} & \cellcolor{blue!20}{$>9 min$} & \cellcolor{green!20} $\sim$0.10 s \\
\midrule

\multirow{4}{*}{\textit{RPR}}

 & ReLoc3r
 & 1.22 / 0.73 & 0.42 / 0.36 & 0.62 / 0.55
 & 0.13 / 0.58 & 0.34 / 0.58 & 0.55 / 0.56 & \cellcolor{green!20}{$< 1min$} & \cellcolor{green!20} 0.23 s\\
 & GeLoc3r
 & 1.07 / 0.52 & 0.41 / 0.34 & 0.64 / 0.54
 & 0.15 / 0.64 & 0.27 / 0.58 & 0.51 / 0.52 & \cellcolor{green!20}{$< 1min$} & \cellcolor{green!20} -- \\
 & \textbf{MultiLoc} (ours)
 & \best{0.45} / \best{0.26} & \best{0.32} / \best{0.32} & \best{0.39} / \best{0.53}
 & \best{0.08} / \best{0.29} & \best{0.23} / \best{0.37} & \best{0.29} / \best{0.35} & \cellcolor{green!20}{$< 1min$} & \cellcolor{green!20} 0.15 s\\
\bottomrule
\end{tabular}%
}
\vspace{-2mm}
\end{table}

\begin{table*}[h]
\centering
\caption{Visual re-localization benchmark on the WaySpots dataset in median translation~/~rotation error as (cm~/~$^\circ$). MASt3R+Kapture is omitted because it's superset, Map-Free, was used to train it.}
\label{tab:wayspots}
\setlength{\tabcolsep}{3pt}
\renewcommand{\arraystretch}{1.3}
\resizebox{\textwidth}{!}{%
\begin{tabular}{ll cccccccccc | c }
\toprule
\rowcolor{headercolor}
\textbf{Categ.} & \textbf{Method} & \textbf{Bears} & \textbf{Cubes} & \textbf{Insc} & \textbf{Lawn} & \textbf{Map} & \textbf{SqB} & \textbf{Statue} & \textbf{Tendrils} & \textbf{Rock} & \textbf{WinterSign} & \textbf{Avg}  \\
\midrule

\multirow{1}{*}{\textit{MC}}

 & E5+1 (RoMa)
 & \best{2.2} / \best{0.72} & \best{8.0} / \best{0.85} & \best{7.9} / \best{1.00} & \best{11.6} / \best{3.29} & \best{7.7} / \best{0.92} & \best{12.3} / \best{0.74} & \best{59.3} / \best{1.48} & \best{22.2} / \best{3.59} & \best{3.1} / \best{0.73} & \best{170.0} / \best{1.55} & \best{30.4} / \best{1.49} \\
\midrule

\multirow{2}{*}{\textit{SCR}}
 & ACE
 & 3.6 / 1.15 & \best{5.3} / \best{0.76} & \best{10.0} / 1.50 & 120.0 / 37.00 & 7.5 / 1.10 & \best{5.7} / \best{0.70} & 595.0 / 15.70 & 166.0 / 45.70 & \best{2.9} / \best{0.74} & \best{423.0} / \best{1.19} & 134.0 / 10.55 \\
 & GLACE
 & \best{3.0} / \best{0.90} & 5.9 / 0.80 & \best{10.0} / \best{1.30} & \best{115.0} / \best{26.30} & \best{7.2} / \best{1.00} & \best{5.7} / \best{0.70} & \best{517.0} / \best{10.30} & \best{155.0} / \best{32.20} & 3.2 / 0.80 & 465.0 / 1.30 & \best{128.7} / \best{7.56}  \\
\midrule

\multirow{2}{*}{\textit{RPR}}

 & ReLoc3r
 & 13.1 / 3.63 & 69.4 / 2.14 & 33.0 / 3.28 & 41.5 / 11.03 & 16.4 / 2.00 & 27.7 / 1.73 & 331.2 / 11.00 & 82.9 / 15.31 & 8.3 / 1.34 & 607.0 / 5.15 & 123.1 / 5.66  \\

 & MultiLoc (ours)
 & \best{4.0} / \best{0.77} & \best{8.3} / \best{0.93} & \best{9.2} / \best{1.09} & \best{16.0} / \best{2.84} & \best{6.9} / \best{1.12} & \best{9.9} / \best{0.75} & \best{89.0} / \best{1.46} & \best{34.0} / \best{3.55} & \best{4.9} / \best{0.72} & \best{65.0} / \best{2.61} & \best{24.7} / \best{1.58} \\
\bottomrule
\end{tabular}%
}
\end{table*}

\subsection{Ablation Study}
\label{subsec:ablation}

\begin{figure}[!h]
  \centering
   \includegraphics[width=0.70\textwidth]{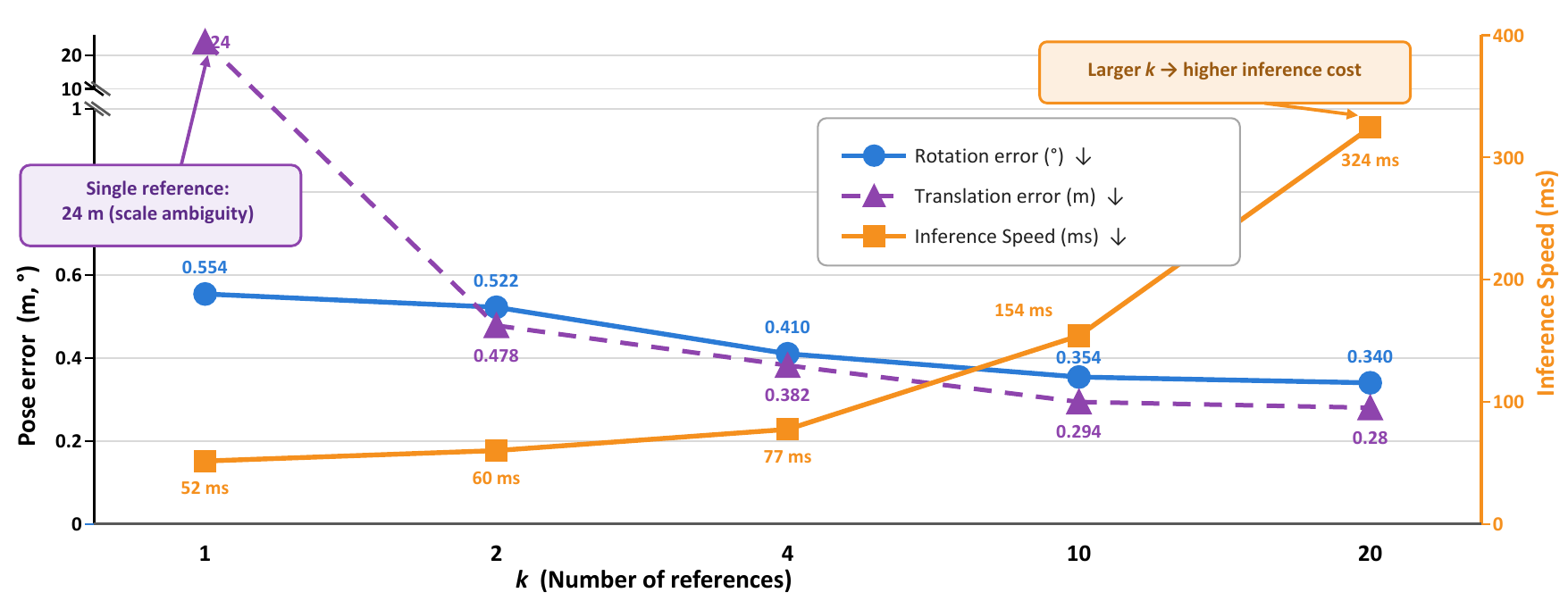}
   \caption{Visual Localization performance (average across all scenes) and inference speed comparison across varying numbers of input views \textit{(k)} on the Cambridge Landmarks.}
   \label{fig:error_vs_views}
\end{figure}

\noindent\textbf{Impact of \textit{k} for localization.}
In this section, we analyze the impact of parameter $k$—representing number of reference images—on both localization accuracy and computational latency. As illustrated in \Cref{fig:error_vs_views}, both translation and rotation error decreases as $k$ increases; however, this gain comes at the cost of higher latency. We observe that $k=10$ provides an optimal balance for MultiLoc, delivering high-precision results while maintaining efficient real-time performance. It is also worth noting that with just two or four retrieved images, our method achieves state-of-the-art RPR localization at 60 and 77 ms, respectively, making it the fastest baseline.

\begin{table}[t]
\centering
\small
\setlength{\tabcolsep}{7pt}
\caption{Median pose error [cm~/~$^\circ$] for visual re-localization (all 5 Cambridge, 10 WaySpots sequences) and AUC@5/10 for camera pose estimation (ACID, ScanNet-1500). \textcolor{cyan}{\faSnowflake}: frozen (no pose intrusion); \textcolor{orange}{\faFire}: trained MultiLoc. \textcolor{red}{Red} denotes best, \textcolor{blue}{blue} second best result.}
\label{tab:ablate_multiloc}
\begin{tabular}{@{}l cc cc@{}}
\toprule
& \multicolumn{2}{c}{\textbf{Visual Reloc.}} & \multicolumn{2}{c}{\textbf{Camera Pose Est.}} \\
\cmidrule(lr){2-3}\cmidrule(lr){4-5}
\textbf{State} & Cambridge & WaySpots & ACID & ScanNet-1500 \\
\midrule
Frozen (no pose)~\textcolor{cyan}{\faSnowflake}
  & \textcolor{blue}{41.0}\,/\,\textcolor{blue}{0.374}
  & \textcolor{blue}{66.6}\,/\,\textcolor{blue}{1.80} & \textcolor{blue}{22.55}\,/\,\textcolor{blue}{40.85} & \textcolor{blue}{38.82}\,/\,\textcolor{blue}{60.31} \\
MultiLoc (with pose)~\textcolor{orange}{\faFire}
  & \textcolor{red}{29.5}\,/\,\textcolor{red}{0.35}
  & \textcolor{red}{24.7}\,/\,\textcolor{red}{1.39} & \textcolor{red}{76.56}\,/\,\textcolor{red}{87.15} & \textcolor{red}{62.34}\,/\,\textcolor{red}{76.32} \\
\bottomrule
\end{tabular}
\end{table}

\noindent\textbf{Effect of training.} We understand the effectiveness of our task-specific training, or developing 3D sub-scene context, by comparing MultiLoc with the frozen image-only variant that is pose free as shown \Cref{tab:ablate_multiloc}. Training consistently improves both re-localization benchmarks. Median translation error falls from $41.0$ to $29.5\,\mathrm{cm}$ on Cambridge ($\sim$28\%) and from $66.6$ to $24.7\,\mathrm{cm}$ on WaySpots ($\sim$63\%). Rotation error improves more modestly, by approximately 6\% and 22\%, respectively. For camera pose estimation, AUC@5/10 increases from $22.55/40.85\%$ to $76.56/87.15\%$ on ACID and from $38.82/60.31\%$ to $62.34/76.32\%$ on ScanNet-1500.

For pose-free visual re-localization, we use VGGT's aggregator and camera weights as the pose regressor, then recover scale with motion averaging, as described in the appendix. These results suggest that the frozen backbone already encodes strong rotational and geometric priors, while task-specific training substantially improves metric translation by calibrating predictions to each scene's scale and layout. MultiLoc thus learns a more accurate, scene-grounded notion of pose, which off-the-shelf geometric features leave underdetermined.

\section{Conclusion and limitations}
\label{sec:Conclusion}

We present MultiLoc, a large-scale, multi-view pose regressor that recasts visual localization as a 3D sub-scene representation-learning problem for pose regression. It's regressor learns to encode a set of reference views together with their camera poses into a single, pose-conditioned scene representation from which the query pose is directly decoded. Such a formulation provides effective geometric grounding, generalization to diverse environments and robust performance competitive to non-pose-regression methods. Nevertheless, performance depends on retrieval quality as with any relative pose regression work. For example, we observe references lacking co-visibility with the query contributes high pose errors, see appendix. Secondly, attending to all image tokens also incurs high latency and memory costs, which token merging or pruning could reduce. Retrieval memory and mapping costs scale with database size; as prior work shows, subsampling redundant frames can reduce them with minimal accuracy loss.

\subsection*{AI use statement}

In this work, we used generative AI tools for paraphrasing and polishing our writing. We have not used generative AI tools for other tasks with required disclosure, and the rest of the required disclosure tasks are not applicable to this work. Additionally, we used generative AI tools for modifying figures. We have reviewed all AI-assisted work. We take responsibility for the final content of this work, including text, claims or artifacts produced with the aid of generative AI.

\subsection*{Ethics statement}

We have no concerns regarding this.

\subsection*{Reproducibility statement}

Our work and the results reported are reproducible. The code for reproducibility will be made public once accepted along with the trained AI model's checkpoint. 

\subsubsection*{Acknowledgments}
This work was supported by Clemson University’s Virtual Prototyping of Autonomy Enabled Ground Systems (VIPR-GS), under Cooperative Agreement W56HZV-21-2-0001 with the US Army DEVCOM Ground Vehicle Systems Center (GVSC).

\bibliography{iclr2027_conference}
\bibliographystyle{iclr2027_conference}

\newpage

\appendix
\section*{Appendix}
\section{Additional Training Configuration}
\label{sec: training_appendix}
 
\subsection{Architecture Ablation}
\label{sec: ablate_appendix}
\begin{figure}[h]
  \centering
   \includegraphics[width=0.94\textwidth]{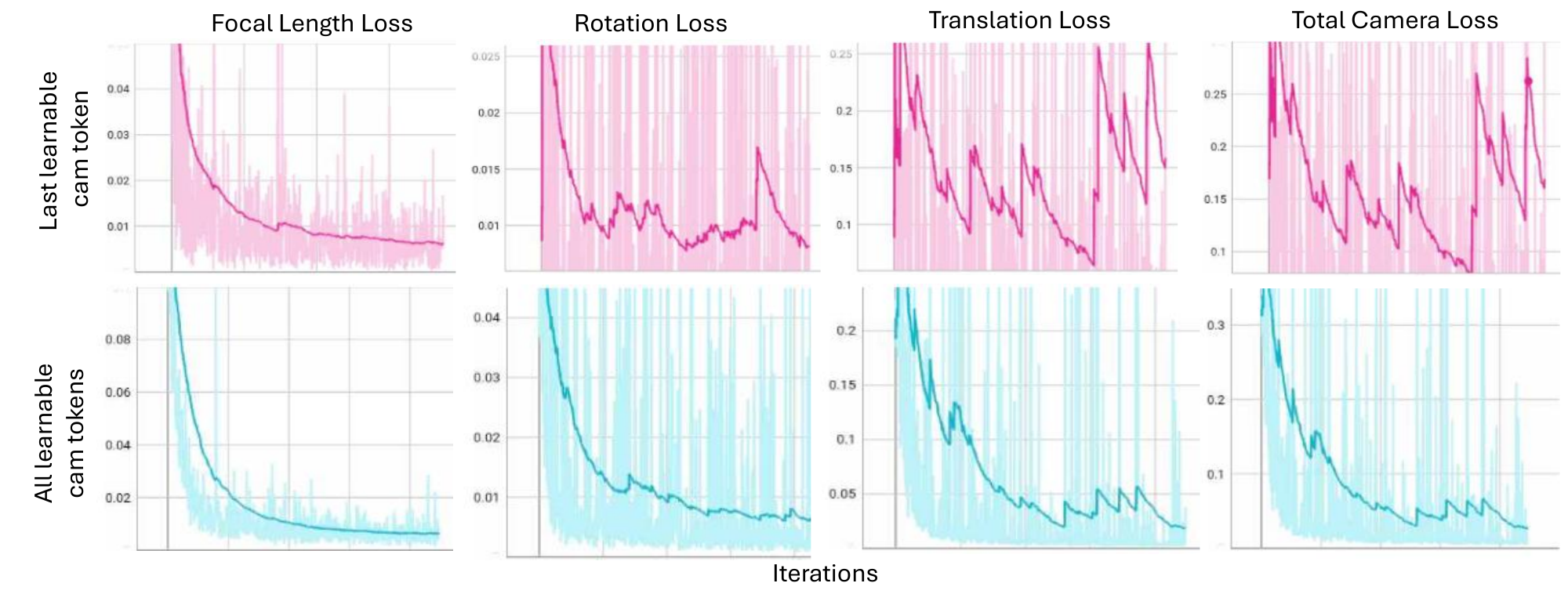}
   \caption{\textbf{Ablation study for MultiLoc's architecture}. Using learnable camera token only for the last image (top) leads to unstable training and sudden gradient jumps. Whereas, using learnable tokens for all the images in a sequence (bottom) and regressing upon them reduces this instability.} 
   \label{fig:arch_ablate}
\end{figure}

MultiLoc's pose regressor employs learnable camera tokens for both reference and query images. While providing learnable tokens for images with known ground truth poses may initially appear redundant, empirical evidence suggests that utilizing a collective set of learnable tokens, $l$, significantly enhances training stability. As illustrated in \cref{fig:arch_ablate}, restricting learnable tokens to the query image alone (i.e., last learnable token) induces volatile gradient spikes, whereas our 'all-learnable' configuration facilitates smoother convergence. A similar holistic token learning design is also employed in concurrent work \cite{peng2025omnivggt}, which also reports reduced error when learnable tokens and ground-truth poses are used together. Due to the high computational overhead of full-scale training and availability of limited compute, these architectural ablations were initially conducted on the ScanNet++ \cite{yeshwanth2023scannet++} dataset as a representative for our broader multi-dataset training.

\section{More Experiments}
\label{sec: experiments_appendix}

\subsection{Scale Recovery}
We highlight the scale recovery of our relative camera poses using three methodologies. The first is a solution to the Procrustes Analysis; Umeyama Alignment \cite{umeyama2002least}, and second one is a motion averaging module that was suggested in \cite{dong2025reloc3r}. Finally, third utilizes the scene scale $s$ that is known from normalizing the camera pose centers of reference images as discussed in the \Cref{subsec:arch}. For the former, to resolve the scale and gauge ambiguity between our model's relative predictions and the global coordinate system, we perform a 7-DoF similarity alignment using the $k$ reference frames. For third, the output pose prediction from our regressor is just multiplied with $s$ to derive the metric pose. We utilize the Umeyama algorithm to compute a transformation $S \in \text{Sim}(3)$ that maps the local relative coordinate frame to the metric world space. This transformation is then applied to the query pose to obtain its final global representation.
For each scene, we report the median translation and rotation errors. As \Cref{tab:scale_recovery} shows, recovering the metric pose with the scale factor $s$---derived directly from the reference images---outperforms both Umeyama alignment and motion averaging. Although motion averaging comes closest, it relies on post-hoc geometric optimization, whereas scale recovery from $s$ requires no such step and is effectively compute-free.

\begin{table*}[ht]
\centering
\caption{Scale Recovery benchmark of MultiLoc (ours) on Cambridge dataset for visual re-localization. Median translation error is in meters (m) and median rotation error in degrees ($^\circ$).}
\label{tab:scale_recovery}
\resizebox{0.98\textwidth}{!}{%
\begin{tabular}{l|ccccc|c|c}
\toprule
Methods & GreatCrt & KingsClg & OldHosp & ShopFac & StMarys & Avg Rot ($^\circ$$\downarrow$) & Avg Trans. (m$ \downarrow$)  \\ \midrule
Umeyama Align. & 0.44 / 0.37 & 0.33 / 0.41 & 0.39 / 0.70 & 0.09 / 0.33 & 0.23 / 0.51 & 0.46 & 0.30 \\
Motion Avg. & 0.45 / 0.25 & 0.32 / 0.33 & 0.40 / 0.52 & 0.08 / 0.28 & 0.24 / 0.36 & \best{0.35} & 0.30 \\
Scale ($s$) & 0.45 / 0.26 & 0.32 / 0.32 & 0.39 / 0.53 & 0.08 / 0.29 & 0.23 / 0.37 & \best{0.35} & \best{0.29} \\
\bottomrule
\end{tabular}%
}
\end{table*}

\subsection{Multi-view for 2D-2D feature matching}
\label{subsec:multiview_matching}
\begin{figure*}[h]
  \centering
   \includegraphics[width=0.68\textwidth]{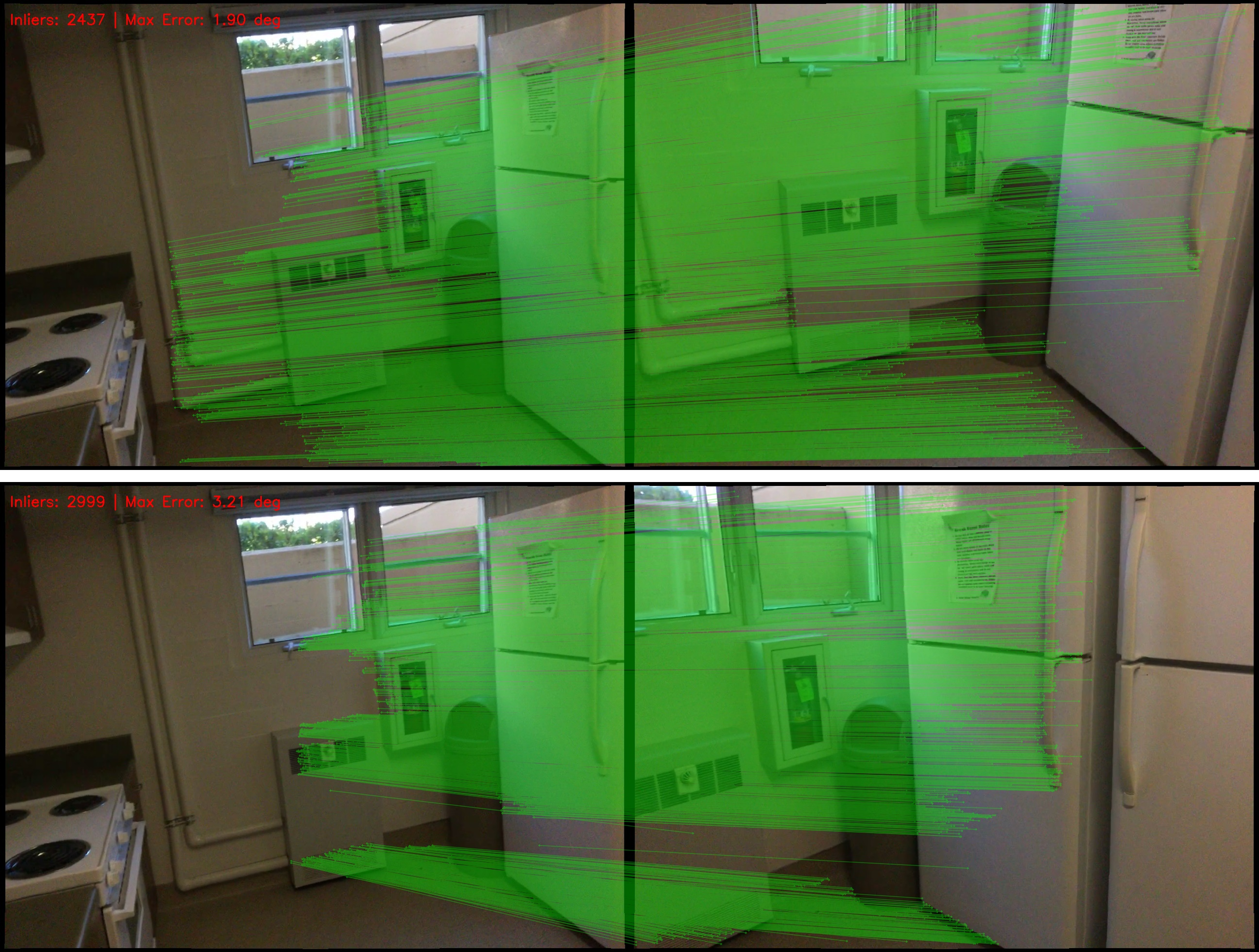}
   \caption{\textbf{RoMA comparison.} Pairwise (top) vs. best-reference (bottom) matching. While additional views increased inliers from 2437 to 2999 (top left in red), pose error, max(trans,rot), rose from 1.90 $\circ$ to 3.21 $\circ$ showing failure to exploit multi-view cues.} 
   \label{fig:roma_multiview}
\end{figure*}

The premise of using multiple views for correspondence based methods, as suggested in \cref{subsec:rel_cam_pose}, is that a support view with a smaller baseline or higher overlap yields more correspondences and, in turn, a better-conditioned pose. This does not hold in practice: the inlier count reflects the quantity and spatial support of correspondences, not the geometric conditioning of the resulting pose. \Cref{fig:roma_multiview} shows a representative case, with error measured as $\max(e_R, e_t)$ (rotation and translation-direction error, in degrees). Matching the query directly to the intended target yields $2437$ inliers and an error of $1.90^\circ$, whereas the highest-inlier support view yields $2999$ inliers yet a \emph{larger} error of $3.21^\circ$. The selector therefore discards the more accurate direct pair in favor of a higher-inlier but geometrically inferior one.

\subsection{Indoor6 results}
\label{subsec:indoor6_full}

In this section we provide the full table for the indoor6 benchmark as shown in \Cref{tab:indoor6_results}. We also provide results for FastForward as discussed in \Cref{sec:intro}.

\begin{table*}[h]
\centering
\caption{Visual re-localization benchmark on the Indoor-6 dataset. We report median translation~/~rotation error (cm~/~$^\circ$); lower is better. The best result \emph{within each category} is in \textbf{bold}. ``--'' denotes results not yet available.}
\label{tab:indoor6_results}
\setlength{\tabcolsep}{5pt}
\renewcommand{\arraystretch}{1.3}
\resizebox{\textwidth}{!}{%
\begin{tabular}{ll cccccc | c }
\toprule
\rowcolor{headercolor}
\textbf{Category} & \textbf{Method} & \textbf{Scene1} & \textbf{Scene2a} & \textbf{Scene3} & \textbf{Scene4a} & \textbf{Scene5} & \textbf{Scene6} & \textbf{Avg} \\
\midrule

\multirow{2}{*}{\textit{MC}}
 & MASt3R + Kapture
 & 3.4 / \best{0.31} & 3.4 / \best{0.15} & 2.9 / \best{0.28} & \best{2.5} / \best{0.39} & 4.2 / \best{0.45} & 2.3 / \best{0.23} & 3.1 / \best{0.30}  \\
 & E5+1 (RoMa)
 & \best{2.2} / 0.43 & \best{2.1} / 0.23 & \best{1.5} / 0.37 & 3.0 / 0.68 & \best{3.8} / 0.65 & \best{1.3} / 0.31 & \best{2.3} / 0.45 \\
\midrule

\multirow{2}{*}{\textit{SCR}}
 & ACE
 & 17.7 / 2.40 & 8.0 / 0.80 & 9.0 / 1.60 & 4.6 / 0.90 & 21.6 / 4.00 & 5.2 / 0.90 & 11.0 / 1.80 \\
 & GLACE
 & \best{3.8} / \best{0.70} & \best{3.9} / \best{0.40} & \best{2.9} / \best{0.60} & \best{2.3} / \best{0.50} & \best{6.1} / \best{0.90} & \best{2.1} / \best{0.40} & \best{3.5} / \best{0.60} \\
 & FastForward
 & -- & -- & -- & -- & -- & -- & 4.0 / \best{0.60} \\
\midrule

\multirow{2}{*}{\textit{RPR}}

 & ReLoc3r
 & 8.0 / 0.60 & 11.0 / 0.78 & 4.0 / 0.44 & 9.0 / 0.87 & 29.0 / 1.60 & 3.0 / 0.52 & 10.7 / 0.80 \\
 & \textbf{MultiLoc} (ours)
 & \best{2.8} / \best{0.41} & \best{2.9} / \best{0.23} & \best{1.9} / \best{0.40} & \best{3.4} / \best{0.62} & \best{4.8} / \best{0.49} & \best{1.6} / \best{0.40} & \best{2.9} / \best{0.43} \\
\bottomrule
\end{tabular}%
}
\end{table*}

\subsection{Cambridge results}
\label{subsec:cambridge_full}
 Full visual re-localization results, including FastForward and HLoc, are provided in \Cref{tab:cambridge_full_res}. Note that, FastForward uses top-20 retreived images for localization on Cambridge benchmark, wheras MultiLoc utilize only 10 while achieving similar accuracy. 

\begin{table}[t]
\centering
\caption{Visual re-localization benchmark on the Cambridge Landmarks dataset in median translation~/~rotation error as (m~/~$^\circ$). Green, blue, and red cells indicate the best, middle, and worst comparative results, respectively.}
\label{tab:cambridge_full_res}
\setlength{\tabcolsep}{5pt}
\renewcommand{\arraystretch}{1.3}
\resizebox{\textwidth}{!}{%
\begin{tabular}{ll ccccc | c | c | c}
\toprule
\rowcolor{headercolor}
\textbf{Category} & \textbf{Method} & \textbf{GreatCourt} & \textbf{Kings} & \textbf{OldHosp} & \textbf{ShopFac} & \textbf{StMary} & \textbf{Avg} \\
\midrule

\multirow{2}{*}{\textit{MC}}
 & HLoc (SP+SG)
 & 0.16 / \best{0.10} & 0.12 / 0.20 & 0.15 / 0.30
 & 0.04 / 0.20 & 0.07 / 0.20 & 0.11 / 0.20  \\
 & MASt3R + Kapture
 & \best{0.10} / \best{0.10} & \best{0.08} / \best{0.12} & \best{0.13} / \best{0.24}
 & \best{0.03} / \best{0.13} & \best{0.04} / \best{0.12} & \best{0.08} / \best{0.14} \\
 & E5+1 (RoMa)
 & 0.37 / 0.16 & 0.21 / 0.35 & 0.29 / 0.64
 & 0.07 / 0.27 & 0.11 / 0.37 & 0.21 / 0.36 \\
\midrule

\multirow{2}{*}{\textit{SCR}}
 & ACE
 & 0.44 / 0.20 & 0.31 / 0.40 & 0.30 / 0.60
 & 0.06 / 0.30 & 0.20 / 0.60 & 0.26 / 0.42  \\
 & GLACE
 & \best{0.19} / \best{0.10} & \best{0.19} / \best{0.30} & \best{0.18} / \best{0.40}
 & \best{0.05} / \best{0.20} & \best{0.10} / \best{0.30} & \best{0.14} / \best{0.26} \\
& FastForward
 & 0.62 / 0.40 & 0.24 / 0.40 & 0.26 / 0.50
 & 0.08 / 0.40 & 0.14 / 0.50 & 0.27 / 0.40 \\
\midrule

\multirow{4}{*}{\textit{RPR}}
 & Map-free (regress)
 & 8.40 / 4.56 & 2.44 / 2.54 & 3.73 / 5.23
 & 0.97 / 3.17 & 2.91 / 5.10 & 3.69 / 4.12  \\
 & ReLoc3r
 & 1.22 / 0.73 & 0.42 / 0.36 & 0.62 / 0.55
 & 0.13 / 0.58 & 0.34 / 0.58 & 0.55 / 0.56 \\
 & GeLoc3r
 & 1.07 / 0.52 & 0.41 / 0.34 & 0.64 / 0.54
 & 0.15 / 0.64 & 0.27 / 0.58 & 0.51 / 0.52  \\
 & \textbf{MultiLoc} (ours)
 & \best{0.45} / \best{0.26} & \best{0.32} / \best{0.32} & \best{0.39} / \best{0.53}
 & \best{0.08} / \best{0.29} & \best{0.23} / \best{0.37} & \best{0.29} / \best{0.35} \\
\bottomrule
\end{tabular}%
}
\end{table}

\subsection{Non-Co-visibility in Retrieval}
\begin{figure*}[h]
  \centering
   \includegraphics[width=0.98\textwidth]{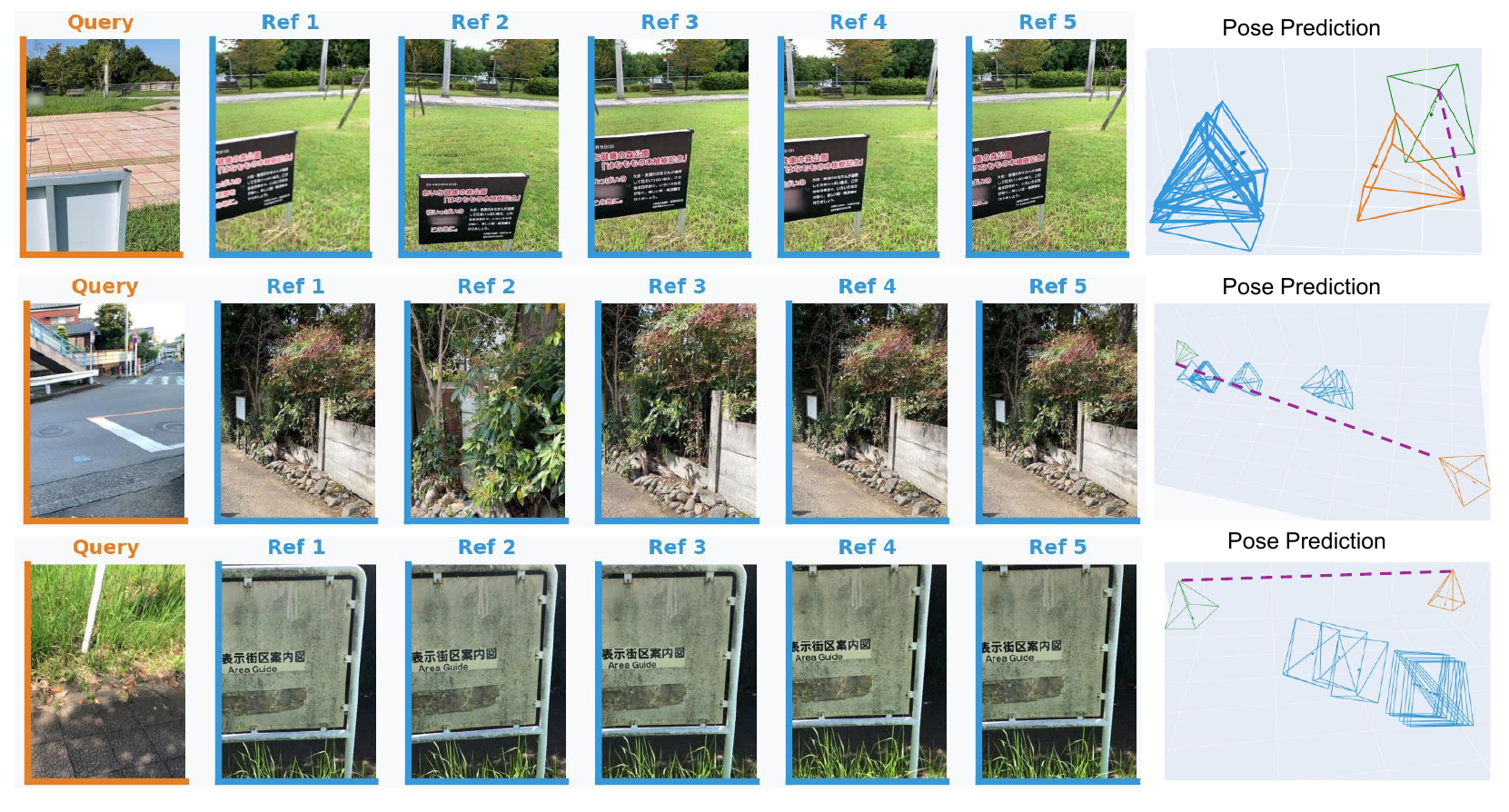}
   \caption{\textbf{Failure cases when retrieved images are not co-visible.} The following cases illustrate scenarios where co-visibility-based retrieval features fail to identify appropriate reference images. We demonstrate using top-5 of 10 such reference images per query. Such instances result in significant translation (\textcolor{purple}{purple dashed line}) and rotation errors between the \textcolor{Goldenrod}{ground truth} and \textcolor{ForestGreen}{predicted} camera poses.} 
   \label{fig:covis_failure}
\end{figure*}

MultiLoc uses NetVLAD as its primary method for image retrieval and reference view selection. We also evaluate it against other retrieval methods, including the co-visibility-driven features introduced in MegaLoc \cite{berton2025megaloc}. In this section, we present several failure cases in which the retrieved images—whether selected by MegaLoc or NetVLAD—are not co-visible with the query, as shown in \Cref{fig:covis_failure}. One of the interesting scenarios from \Cref{fig:covis_failure} highlights the street viewing direction that fails to successfully retrieve reference images. Furthermore, the exact scenarios where MegaLoc performance is suboptimal is mentioned in \cite{berton2025megaloc}. Given the rapid advancements in feed-forward 3D geometry models, optimizing reference view selection with co-visibility criterion remains a critical challenge, which we designate for future works.

\section{Details about baselines}
\label{sec: details_baselines}
For every baseline we follow the authors' original public implementations and default settings, and we take care to place all methods on a comparable footing. For MASt3R \cite{leroy2024grounding}, input images are resized so that the longer side is 512 and then center-cropped to a multiple of 16 (model's patch constraint), with the intrinsics adjusted for both operations. For RoMa \cite{edstedt2024roma}, different from ReLoc3r \cite{dong2025reloc3r} and NoPoSplat's \cite{ye2024no} implementation, we adopt RoMa's original MegaDepth benchmarking pipeline, which accepts an arbitrary input size, resizes to 560×560 resolution, and finally upsamples 864×864 resolution. Sparse correspondences are then mapped back to a per-dataset pixel frame in which the RANSAC threshold is interpreted: the longer side is set to 1200 px for MegaDepth-1500, the shorter side to 480 px for ScanNet-1500 following DKM/RoMa convention, and native resolution is used for ACID. We use the outdoor RoMa weights for MegaDepth-1500, ACID, Cambridge, and WaySpots, and the indoor weights for ScanNet-1500 and Indoor6 benchmarks.

For the two-view relative-pose benchmarks, both RoMa and MASt3R estimate the pose from 2D–2D correspondences via an essential matrix, so that MASt3R's metric point maps are not used and the comparison is purely a matching comparison. From RoMa's dense field we sample 5000 correspondences per pair. MASt3R contributes up to 12000 reciprocal-nearest-neighbour matches. In both cases the RANSAC inlier threshold is a 0.5 px error normalized by the mean focal length. This full-image essential-matrix setting reflects each matcher's strongest two-view configuration, since the feature matcher observes the entire image rather than a cropped region.

We further evaluate MASt3R in a structure-based localization setting as denoted in \Cref{sec:experiment} as MASt3r+Kapture. Given a 3D map of posed database images, the query is matched to the map, and its absolute pose is recovered by PnP using PoseLib with a reprojection threshold of 0.008× the image diagonal.

For ReLoc3r, we follow its resize-and-center-crop protocol where the longer side is 512 and shorter side often being 384 pixels. For VGGT \cite{wang2025vggt} and MultiLoc we resize the longest dimension to be 518 pixels based on VGGT's implementation. Finally, for the scene-coordinate-regression methods ACE \cite{brachmann2023accelerated} and GLACE \cite{wang2024glace} we use their original implementations and recommended settings.

\end{document}